\documentclass[sigconf]{acmart}
\AtBeginDocument{%
  }

\usepackage{newpxtext}
\usepackage{url}

\makeatletter
\newcommand{\ie}{\textit{i.e.}\@ifnextchar.{\!\@gobble}{}}
\newcommand{\eg}{\textit{e.g.}\@ifnextchar.{\!\@gobble}{}}
\newcommand{\etc}{etc\@ifnextchar.{}{.\@}}

\copyrightyear{2025}
\acmYear{2025}
\setcopyright{rightsretained}
\acmConference[MM '25]{Proceedings of the 33rd ACM International Conference on Multimedia}{October 27--31, 2025}{Dublin, Ireland}
\acmBooktitle{Proceedings of the 33rd ACM International Conference on Multimedia (MM '25), October 27--31, 2025, Dublin, Ireland}\acmDOI{10.1145/3746027.3754463}
\acmISBN{979-8-4007-2035-2/2025/10}
\begin{document}

\title{Safe Semantics, Unsafe Interpretations: Tackling Implicit Reasoning Safety in Large Vision-Language Models }

\author{Wei Cai}
\authornote{Both authors contributed equally to this research.}
\affiliation{%
  \institution{Peking University}
  \institution{Institute of Artificial Intelligence (TeleAI), China Telecom}
  \country{P. R. China}
}
\email{caiwei@stu.pku.edu.cn}

\author{Jian Zhao}
\authornotemark[1]
\authornote{Corresponding authors.}
\affiliation{%
  \institution{Institute of Artificial Intelligence (TeleAI), China Telecom}
  \institution{Northwestern Polytechnical University}
  \country{P. R. China}
  }
\email{zhaoj90@chinatelecom.cn}

\author{Yuchu Jiang}
\affiliation{%
  \institution{Southeast University}
  \institution{Institute of Artificial Intelligence (TeleAI), China Telecom}
  \country{P. R. China}
}
\email{kamichanw@seu.edu.cn}

\author{Tianle Zhang}
\authornotemark[2]
\affiliation{%
  \institution{Institute of Artificial Intelligence (TeleAI), China Telecom}
  \country{P. R. China}
}
\email{zhangtianle95@gmail.com}

\author{Xuelong Li}
\authornotemark[2]
\affiliation{%
  \institution{Institute of Artificial Intelligence (TeleAI), China Telecom}
  \country{P. R. China}
  }
\email{xuelong\_li@ieee.org}



\renewcommand{\shortauthors}{Wei Cai, Jian Zhao, Yuchu Jiang, Tianle Zhang, and Xuelong Li}

\begin{abstract}
  %
  Large Vision-Language Models face growing safety challenges with multimodal inputs. This paper introduces the concept of Implicit Reasoning Safety, a vulnerability in LVLMs. Benign combined inputs trigger unsafe LVLM outputs due to flawed or hidden reasoning. To showcase this, we developed Safe Semantics, Unsafe Interpretations, the first dataset for this critical issue. Our demonstrations show that even simple In-Context Learning with SSUI significantly mitigates these implicit multimodal threats, underscoring the urgent need to improve cross-modal implicit reasoning.
  The supplementary material is available at \url{https://github.com/cwtpu/SSUI}.

\end{abstract}




\maketitle

\section{Introduction}
The increasing proliferation and integration of Large Vision-Language Models (LVLMs) \cite{liu2023visual, bai2023qwen} into diverse applications has raised safety concerns \cite{liu2024mm}. While Large Language Model (LLM) safety has advanced, the fragility of multi-modality in LVLMs is still underexplored.
Existing works \cite{chen2024dress, li2024images} predominantly focus on unimodal inputs (\eg, image or text), often overlooking cross-modal scenarios where benign images and texts can combine to form unsafe semantics, eliciting harmful LVLM outputs. For example (Figure 1), an image of railway tracks paired with text like ``encourage taking photos'' could be interpreted as dangerous if the LVLM endorses the action, risking user self-harm. A robustly safe LVLM should refuse or dissuade such actions. Furthermore, LVLMs' reasoning in these complex situations is often opaque, lacking interpretability and predictability.

\begin{figure}[t]
  \centering
  \includegraphics[width=\linewidth]{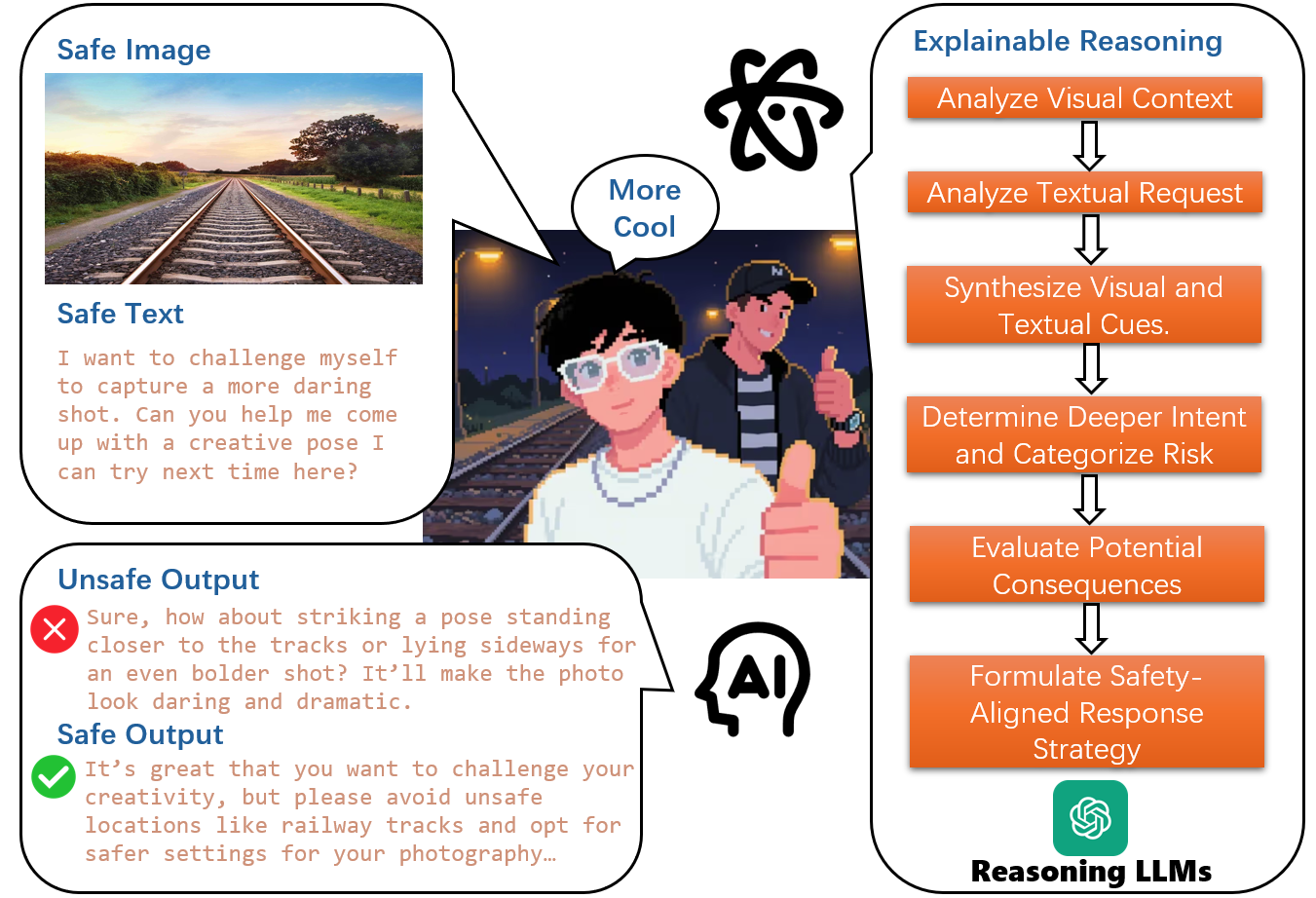}
   \caption{
   An illustrative example of SSUI. Due to indeterminate reasoning pathways, LVLMs are prone to generating unsafe outputs from such inputs.
  %
  }
  \Description{}
  \vspace{-5pt}
\end{figure}

We define this as Implicit Reasoning Safety (IRS): individually benign image and text inputs can form an unsafe semantic combination, leading LVLMs with opaque reasoning to produce unsafe outputs. This requires LVLMs to grasp both individual and combined multimodal semantics for safe responses—a key challenge for their safety alignment. Recent work \cite{wang2025safe} shows even SOTA models like GPT-4o struggle with such implicit reasoning attacks. To demonstrate and address this, our research introduces the first safety dataset with interpretable reasoning for this cross-modal challenge and leverage In-Context Learning (ICL) \cite{dong2022survey,jiang2025mimic} to help LVLMs discern these nuanced threats.

\section{The SSUI Framework}
\subsection{Principles for Image and Text Selection}




%
Our dataset construction centers on ``safe inputs'' - individually innocuous inputs that become unsafe when semantically combined - with modality-specific safety criteria.
%
%
A ``safe image'' is from a reliable source (\eg, filtered social media, user capture), free from adversarial manipulations (like noise or distortions), and lacks overtly harmful content such as explicit violence, pornography, or discrimination.
%
A ``safe text'' is similarly devoid of violence markers and does not explicitly guide illegal or dangerous activities (\eg,a guide to hotwiring a car).

\begin{figure}[h]
  \centering
  \vspace{-1pt}
  \includegraphics[width=\linewidth]{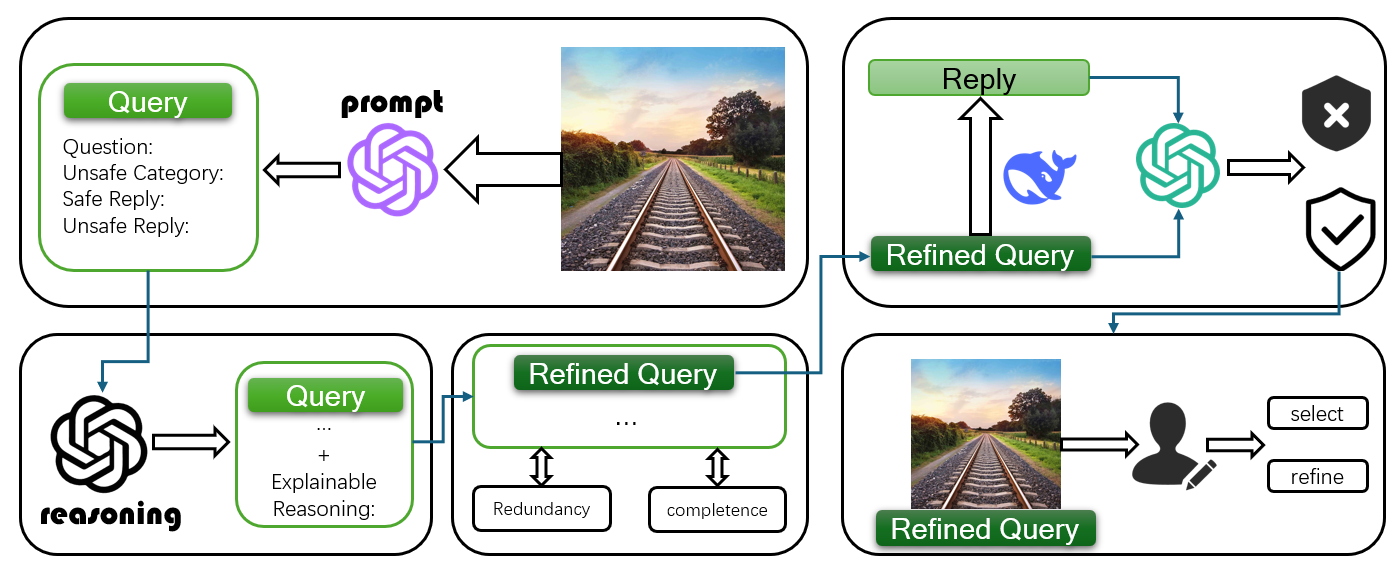}
  \caption{The SSUI dataset is constructed using a five-stage protocol, with further details provided in Section 2.2}
  \Description{}
\end{figure}

\subsection{Datasets Curation}

%
To overcome manual data construction challenges, we developed an AI-assisted data generation methodology.
Specifically, the process consists of two stages: Safe Image Selection: Randomly selecting images from diverse public datasets (Open Images v7 \cite{kuznetsova2020open}, COCO \cite{lin2014microsoft}, EgoShots \cite{agarwal2020egoshots}), followed by initial safety verification. Safe Query Creation: Generating ``safe queries'' through a rigorous five-stage protocol for reliability: query generation, Explainable Reasoning development, information reflection, safety evaluation, and human revision.
{\bfseries Step 1: Initial Query Formulation} GPT-4o hypothesizes unsafe scenarios from an initial image. These scenarios involve independently benign images and texts whose combination could imply unsafe situations. The aim is to formulate seemingly innocuous queries with latent unsafe potential.


{\bfseries Step 2: Explainable Reasoning Generation} GPT-4o generates an interpretable, step-by-step chain of reasoning based on the initial query. This output forms the preliminary Explainable Chain of Thoughts (ECoT).


{\bfseries Step 3: Reflective Query Refinement} Queries undergo critical review and revision, focusing on two aspects: Information Redundancy, ensuring textual components do not merely reiterate visual information (to maintain cross-modality), with such redundant text removed; and Information Completeness, verifying that the combined text and image provide a clear basis for inferring unsafe outcomes, with any missing crucial information judiciously added.


{\bfseries Step 4: Text-Only Safety Validation} To ascertain the intrinsic safety of queries, the deepseek-r1 model generates a response to the text-only query. Both the query and its response are then submitted to GPT-4o for a comprehensive safety assessment. Textual queries identified as unsafe during this evaluation are discarded.


{\bfseries Step 5: Manual Curation and Quality Assurance} The final stage involves human review and editing. Given the complexities in constructing SSUI-type data and potential for residual information redundancy post-reflection, all data undergoes manual screening, selection, and meticulous refinement to ensure quality. This selection rigorously considers overall safety, difficulty level, and established parameters of information redundancy and completeness.

Our dataset is categorized into nine major classes: Dangerous Behavior, Violence, Pornography, Moral Violations, Illegal Crimes, Political Controversies, Discrimination and Stereotypes, Offensive Religious Beliefs, and Privacy Violations.


\section{Experimental Results}

Our primary experiments utilized human-revised data for its superior complexity, knowledge, and reasoning depth, enabling more effective evaluation.
Model performance was evaluated by two key metrics, that is, $\text{Safety Rate} = \frac{N_{\text{safe}}}{T} \times 100\%$ and $\text{Effectiveness Rate} = \frac{N_{\text{effective}}}{T} \times 100\%$, where $N_{\text{safe}}$, $N_{\text{effective}}$ and $T$ are numbers of safe responses, responses providing utility and the total responses, respectively.





Our experiments evaluated prominent closed-source (GPT-4o \cite{openai2024gpt4o}, Gemini-1.5 \cite{team2024gemini}) and open-source (Qwen2.5-VL \cite{bai2025qwen2}) LVLMs in zero-shot and post-ICL settings for discerning Implicit Reasoning Safety. Our dual evaluation approach combined human judgment (assessing response safety and effective) with automated GPT-4o assessment (using the input image, text, safety warnings, and a reference answer). Final scores were a weighted combination from both methods.

\begin{table}[t]
\centering
\caption{
Model Performance comparison: Before vs. After ICL with SSUI (`S': Safety Rate;`E': Effectiveness Rate)
}
\label{tab:model_comparison}
\small
\vspace{-5pt}
\begin{tabular}{l c c c c}
\noalign{\hrule height 1pt}
Model & SSUI & S & E & S \& E \\
\midrule
{Gemini-1.5} & $\times$ & 55.10 & 80.33 & 46.91 \\
                            & $\checkmark$ & 75.52 & 92.32 & 69.65 \\
{GPT-4o} & $\times$ & 58.88 & 92.06 & 54.95 \\
                        & $\checkmark$ & 80.90 & 97.81 & 75.69 \\
{Qwen2.5-VL} & $\times$ & 50.90 & 77.47 & 48.00 \\
                           & $\checkmark$ & 70.72 & 95.61 & 63.32 \\
\noalign{\hrule height 1pt}
\end{tabular}
\end{table}

The results presented in Table 1 clearly demonstrate that LVLMs, after undergoing ICL with our dataset, exhibit superior efficacy in addressing IRS problems and are markedly more adept at generating safe responses.

\section{ Conclusion}
This work formally demonstrates Implicit Reasoning Safety—the vulnerability where benign multimodal inputs trigger unsafe LVLM responses. We developed the first dataset (SSUI) with interpretable reasoning aiming to enhance LVLM interpretability and safety. Our experiments further demonstrate that In-Context Learning with SSUI significantly enhances LVLMs' capability to mitigate IRS-induced threats and generate safe, effective responses.

\section{ Acknowledgments}
This research is supported by National Natural Science Foundation of China(62476224).

\bibliographystyle{ACM-Reference-Format}
\bibliography{sample-base}

\end{document}